\documentclass[9pt,twocolumn,twoside]{osajnl}

\usepackage{times}
\usepackage{epsfig}
\usepackage{graphicx}
\usepackage{float}
\usepackage{wrapfig}

\usepackage{amsmath,amssymb}

\usepackage{bm,xspace}
\usepackage{comment}
\usepackage{verbatim}
\usepackage{multirow}
\usepackage{balance}
\usepackage{url}
\usepackage{booktabs}
\usepackage{etoolbox,siunitx}
\usepackage{calc}
\usepackage{pifont,hologo}
\usepackage{color}

\usepackage[super]{nth}
\usepackage{nicefrac}
\def\aa{\mathbf{a}}

\def\ss{\mathbf{s}}
\def\tt{\mathbf{t}}

\def\AA{\mathbf{A}}

\DeclareMathSymbol{@}{\mathord}{letters}{"3B}

\newcommand\mypara[1]{\vspace{3pt}\noindent{\small\sf \textbf{#1.}}}

\def\latex/{\LaTeX}
\def\bibtex/{\hologo{BibTeX}}

\usepackage[official]{eurosym}

\newcommand{\hide}[1]{}

\usepackage{amsmath} 
\usepackage{amssymb}
\usepackage{bm}
\usepackage{acro}
\usepackage{balance}
\usepackage{booktabs}
\usepackage{pifont}
\usepackage{color}
\usepackage[]{graphicx} 
\usepackage{hyperref}
\usepackage{siunitx}    
\usepackage{times}
\usepackage{tabularx} 
\usepackage{subcaption}
\usepackage{xcolor}
\usepackage{enumitem}

\newcommand{\cmark}{\ding{51}}%
\newcommand{\xmark}{\ding{55}}%

\usepackage[TS1,T1]{fontenc}
\DeclareSIUnit[number-unit-product = {}]{\inch}{\textquotedbl}

\usepackage{tikz,pgfplots}
\usetikzlibrary{matrix}
\pgfplotsset{compat=1.15}
\usepackage{etex} 
\usetikzlibrary{arrows,automata}
\usetikzlibrary{positioning}
\usepgfplotslibrary{groupplots}
\usepackage{tikz-3dplot}
\tikzset{
	state/.style={
		rectangle,
		rounded corners,
		draw=black, very thick,
		minimum height=2em,
		inner sep=2pt,
		text centered,
	},
}
\tikzset{
	info/.style={
		rectangle,
		draw=black, thin,
		minimum height=2em,
		inner sep=2pt,
		text centered,
	},
}
\usetikzlibrary{positioning}
\usetikzlibrary{calc}
\usetikzlibrary{arrows,shapes,backgrounds}
\usepackage{nameref} 

\newcolumntype{Y}{>{\centering\arraybackslash}X}

\newcommand{\rebuttal}[1]{#1}

\newlist{todolist}{itemize}{2}
\setlist[todolist]{label=$\square$}

\newcolumntype{C}[1]{>{\centering\let\newline\\\arraybackslash\hspace{0pt}}m{#1}}

\DeclareAcronym{TWR}{
    short = thrust-to-weight,
    long = thrust-to-weight ratio
}

\DeclareAcronym{TIR}{
    short = torque-to-inertia,
    long = torque-to-inertia ratio
}

\DeclareAcronym{SBC}{
    short = SBC,
    long = single-board computer
}

\DeclareAcronym{ROS}{
    short = ROS,
    long = Robot Operating System
}

\DeclareAcronym{FPV}{
    short = FPV,
    long = first-person-view
}

\DeclareAcronym{HIL}{
    short = HIL,
    long = hardware-in-the-loop
}

\DeclareAcronym{MAV}{
    short = MAV,
    long = micro-aerial vehicle
}

\DeclareAcronym{ASIC}{
    short = ASIC,
    long = application-specific integrated circuit
}

\DeclareAcronym{MPC}{
    short = MPC,
    long = model-predictive control
}

\DeclareAcronym{ISP}{
    short = ISP,
    long = image-signal processor
}

\DeclareAcronym{VIO}{
    short = VIO,
    long = visual-inertial odometry
}

\DeclareAcronym{IMU}{
    short = IMU,
    long = inertial measurement unit
}

\DeclareAcronym{INDI}{
    short = INDI,
    long = incremental non-linear dynamic inversion
}

\DeclareAcronym{RMSE}{
    short = RMSE,
    long = root-mean-square error
}

\DeclareAcronym{ATE}{
    short = ATE,
    long = absolute tracking error
}

\DeclareAcronym{BEM}{
    short = BEM,
    long = blade-element momentum theory
}

\DeclareSIUnit\baud{Baud}
\DeclareSIUnit\flops{FLOPS}
\DeclareSIUnit\g{g}
\DeclareSIUnit\inch{inch}

\newcommand{\agilow}{agiNuttx}
\journal{ol} 

\setboolean{shortarticle}{false}

\title{Agilicious: Open-Source and Open-Hardware Agile Quadrotor for Vision-Based Flight}

\author[$\dagger$,*]{Philipp Foehn}
\author[$\dagger$,*]{Elia Kaufmann}
\author[$ $]{Angel Romero}
\author[$ $]{Robert Penicka}
\author[$ $]{Sihao Sun}
\author[$ $]{Leonard Bauersfeld}
\author[$ $]{Thomas Laengle}
\author[$ $]{Giovanni Cioffi}
\author[$ $]{Yunlong Song}
\author[$ $]{Antonio Loquercio}
\author[$ $]{\\Davide Scaramuzza}

\affil[$ $]{All authors are with the Robotics and Perception Group, UZH, Zurich, Switzerland.}
\affil[$\dagger$]{These authors contributed equally to this work.}
\affil[*]{Corresponding authors: foehn@ifi.uzh.ch, ekaufmann@ifi.uzh.ch}

\dates{This is the accepted version of Science Robotics Vol. 7, Issue 67 \\ DOI: 10.1126/scirobotics.abl6259 (2022) }

\begin{abstract}
Autonomous, agile quadrotor flight raises fundamental challenges for robotics research in terms of perception, planning, learning, and control.
A versatile and standardized platform is needed to accelerate research and let practitioners focus on the core problems.
To this end, we present Agilicious, a co-designed hardware and software framework tailored to autonomous,  agile quadrotor flight.
It is completely open-source and open-hardware and supports both model-based and neural-network--based controllers. 
Also, it provides high thrust-to-weight and torque-to-inertia ratios for agility, onboard vision sensors, GPU-accelerated compute hardware for real-time perception and neural-network inference, a real-time flight controller, and a versatile software stack.
In contrast to existing frameworks, Agilicious offers a unique combination of flexible software stack and high-performance hardware.
We compare Agilicious with prior works and demonstrate it on different agile tasks, using both model-based and neural-network--based controllers. Our demonstrators include trajectory tracking at up to \SI{5}{\g} and \SI{70}{\kilo\meter\per\hour} in a motion-capture system, and vision-based acrobatic flight and obstacle avoidance in both structured and unstructured environments using solely onboard perception. Finally, we  demonstrate its use for hardware-in-the-loop simulation in virtual-reality environments.
Thanks to its versatility, we believe that Agilicious supports the next generation of scientific and industrial quadrotor research.
\end{abstract}

\begin{document}

\maketitle

\section*{Code and Multimedia Material}
Code and data can be found at {\small\url{https://agilicious.dev}}. 
A video of the experiments can be found at {\small\url{https://youtu.be/fNYxPLyJ5YY}}.

\section{Introduction} \label{sec:introduction}

Quadrotors are extremely agile vehicles. Exploiting their agility in combination with full autonomy is crucial for time-critical missions, such as search and rescue, aerial delivery, and even flying cars.
For this reason, over the past decade, research on autonomous, agile quadrotor flight 
has continually pushed platforms to higher levels of speed and agility~\cite{Mellinger2012ijrr, loianno2016estimation,kaufmann18icra, Mohta18jfr, Zhou19arxiv,  Foehn20rss, croonRAS20,Nguyen2020arxiv,kaufmann2020RSS, Foehn2021science}.

To further advance the field, several competitions have been organized---such as the autonomous drone racing series at the recent IROS and NeurIPS conferences~\cite{moon2019challenges,cocoma2019towards,kaufmann18icra,madaan2020airsim} and the AlphaPilot challenge~\cite{guerra2019flightgoggles,Foehn20rss}---with the goal to develop autonomous systems that will eventually outperform expert human pilots.
Million-dollar projects, such as AgileFlight~\cite{AgileFlight} and Fast Lightweight Autonomy (FLA)~\cite{Mohta18jfr}, have also been funded by the European Research Council and the United States government, respectively, to further push research.
Agile flight comes with ever-increasing engineering challenges since performing faster maneuvers with an autonomous system requires more capable algorithms, specialized hardware, and proficiency in system integration. 
As a result, only a small number of research groups have undertaken the significant overhead of hardware and software engineering,
and have developed the expertise and resources to design quadrotor platforms that fulfill the requirements on weight, sensing, and computational budget necessary for autonomous agile flight.
This work aims to bridge this gap through an open-source agile flight platform, enabling everyone to work on agile autonomy with minimal engineering overhead.

\begin{figure*}[t!]
    \centering
    \includegraphics[width=1\linewidth]{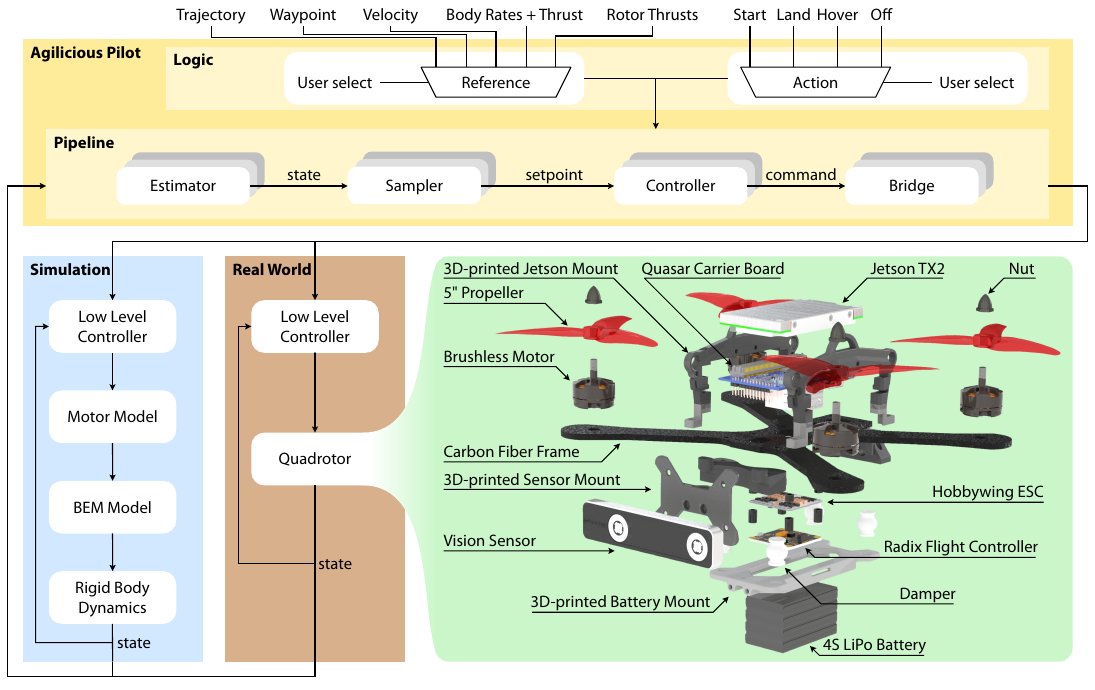}
    \vspace*{-6pt}
    \caption{
    The Agilicious software and hardware quadrotor platform are tailored for agile flight while featuring powerful onboard compute capabilities through an NVIDIA Jetson TX2.
    \rebuttal{
    The versatile sensor mount allows for rapid prototyping with a wide set of monocular or stereo camera sensors.}
    As a key feature, the software of Agilicious is built in a modular fashion, allowing rapid software prototyping in simulation and seamless transition to real-world experiments.
    The Agilicious Pilot encapsulates all logic required for agile flight, while exposing a rich set of interfaces to the user, from high-level pose commands to direct motor commands.
    The software stack can be used in conjunction with a custom modular simulator
    which supports highly accurate aerodynamics based on blade-element momentum theory \cite{bauersfeld2021neurobem}, or with RotorS \cite{Furrer2016rotors}, hardware-in-the-loop, and rendering engines such as Flightmare \cite{yunlong2020flightmare}.
    Deployment on the physical platform only requires selecting a different bridge and a sensor-compatible estimator.
    }
    \label{fig:overview}
\end{figure*}

The platforms and software stacks developed by research groups~\cite{Meier2012pixhawk, loianno2016estimation, Mohta18jfr, Sa2018ram, Faessler18ral, Oleynikova2020jfr, Baca2021jirs, paparazzi}
vary strongly in their choice of hardware and software tools.
This is expected, as optimizing a robot with respect to different tasks based on individual experience in a closed-source research environment leads to a fragmentation of the research community.
For example, even though many research groups use the \acl{ROS} middleware to accelerate development, publications are often difficult to reproduce or verify since they build on a plethora of previous implementations of the authoring research group.
In the worst case, building on an imperfect or even faulty closed-source foundation can lead to wrong or non-reproducible conclusions, slowing down research progress.
To break this vicious cycle and to democratize research on fast autonomous flight, the robotics community needs an open-source and open-hardware quadrotor platform that provides the versatility and performance needed for a wide range of agile flight tasks.
Such an open and agile platform does not yet exist, which is why we present Agilicious, an open-source and open-hardware agile quadrotor flight stack summarized in Figure~\ref{fig:overview}.

To reach the goal of creating an agile, autonomous, and versatile quadrotor research platform, two main design requirements must be met by the quadrotor:
it must carry the required compute hardware needed for autonomous operation, 
and it must be capable of agile flight.

To meet the first requirement on computing resources needed for true autonomy, a quadrotor should carry sufficient compute capability to concurrently run estimation, planning, and control algorithms onboard. 
With the emergence of learning-based methods, also efficient hardware acceleration for neural network inference is required. 
To enable agile flight, the platform must deliver an adequate \acl{TWR} and \acl{TIR}.
While the \acl{TWR} can often be enhanced using more powerful motors, which in turn require larger propellers and thus a larger size of the platform.
However, the \acl{TIR} typically decreases with higher weight and size, since the moment of inertia increases quadratic with the size, and linearly with the weight.
As a result, it is desirable to design a lightweight and small platform~\cite{kumar2012opportunities,Floreano2015nature} to maximize agility (i.e. maximize both \ac{TWR} and \ac{TIR}).
Therefore the platform should meet the best trade-off, 
since maximizing compute resources competes against maximizing the flight performance.

Apart from hardware design considerations, a quadrotor research platform needs to provide the software framework for flexible usage and reproducible research. 
This entails the abstraction of hardware interfaces and a general co-design of software and hardware necessary to exploit the platform's full potential.
Such co-design must account for the capabilities and limitations of each system component, such as the complementary real-time capabilities of common operating systems and embedded systems, communication latencies and bandwidths, system dynamics bandwidth limitations, and efficient usage of hardware accelerators. 
In addition to optimally using hardware resources, the software should be built in a modular fashion to enable rapid prototyping through a simple exchange of components, both in simulation and real-world applications. 
This modularity enables researchers to test and experiment with their research code, without the requirement to develop an entire flight stack, accelerating time to development and facilitating reproducibility of results. 
Finally, the software stack should run on a broad set of computing boards, be efficient, easy to transfer and adapt by having minimal dependencies and provide known interfaces, such as the widely-used Robot Operating System (ROS).

The complex set of constraints and design objectives is difficult to meet.
There exists a variety of previously published open-source research platforms, which, while well designed for low-agility tasks, could only satisfy a subset of the aforementioned hardware and software constraints.
In the following section, we list and analyze prominent examples such as the FLA~platform~\cite{Mohta18jfr}, the MRS~quadrotor~\cite{Baca2021jirs}, the ASL-Flight~\cite{Sa2018ram}, the MIT-Quad~\cite{tal2020tcst}, the GRASP-Quad~\cite{loianno2016estimation}, or our previous work~\cite{Faessler18ral}.

The FLA~platform~\cite{Mohta18jfr} relies on many sensors, including Lidars and laser-range finders in conjunction with a powerful onboard computer. 
While this platform can easily meet autonomous flight computation and sensing requirements, it does not allow to perform agile flight beyond 2.4g of thrust, limiting the flight envelope to near-hover flight. 
The MRS~platform~\cite{Baca2021jirs} provides an accompanying software stack and features a variety of sensors. 
Even though this hardware and software solution allows fully autonomous flight, the actuation renders the system not agile with a maximum thrust-to-weight of 2.5.
The ASL-Flight~\cite{Sa2018ram} is built on the DJI Matrice 100 platform and features an Intel NUC as the main compute resource. 
Similarly to the MRS platform, the ASL-Flight has very limited agility due to its weight being on the edge of the platform's takeoff capability.
The comparably smaller GRASP-Quad proposed in~\cite{loianno2016estimation} operates with only onboard \ac{IMU} and monocular camera while having a weight of only \SI{250}{\gram}.
Nevertheless, the Qualcomm Snapdragon board installed on this platform lacks computational power and also the actuation constrains the maximal accelerations below 1.5g.
Motivated by drone racing, the MIT-Quad~\cite{tal2020tcst} reported accelerations of up to 2.1g while it was further equipped with NVIDIA Jetson TX2 in~\cite{antonini202IJRR}, however, it does not reach the agility of Agilicious and contains proprietary electronics.
Finally, the quadrotor proposed in~\cite{Faessler18ral} is a research platform designed explicitly for agile flight.
Although the quadrotor featured a high thrust-to-weight ratio of up to 4, its compute resources are very limited, prohibiting truly autonomous operation. 
All these platforms are optimized for either relatively heavy sensor setups or for agile flight in non-autonomous settings.
While the former platforms lack the required actuation power to push the state of the art in autonomous agile flight, the latter have insufficient compute resources to achieve true autonomy.

Finally, several mentioned platforms rely on either Pixhawk-PX4~\cite{meier2015px4}, the Parrot~\cite{ParrotANAFI_AI} or DJI~\cite{DJIFPV} low-level controllers, which are mostly treated as blackboxes.
This, together with the proprietary nature of the DJI systems, limits control over the low-level flight characteristics, which not only limits interpretability of results, but also negatively impacts agility.
Full control over the complete pipeline is necessary to truly understand aerodynamic and high-frequency effects, model and control them, and exploit the platform to its full potential.

Apart from platforms mainly developed by research labs, several quadrotor designs are proposed by industry (Skydio~\cite{Skydio}, DJI~\cite{DJIFPV}, Parrot~\cite{ParrotANAFI_AI}) and open-source projects (PX4~\cite{meier2015px4}, Paparazzi~\cite{paparazzi}, Crazyflie~\cite{GiernackiCrazy}). 
While Skydio~\cite{Skydio} and DJI~\cite{DJIFPV} both develop platforms featuring a high level of autonomy, they do not support interfacing with custom code and therefore are of limited value for research and development purposes. 
Parrot~\cite{ParrotANAFI_AI} provides a set of quadrotor platforms tailored for inspection and surveillance tasks that are accompanied by limited software development kits that allow researchers to program custom flight missions. 
In contrast, PX4~\cite{meier2015px4} provides an entire ecosystem of open-source software and hardware as well as simulation. 
While these features are extremely valuable especially for low-speed flight, both cross-platform hardware and software are not suited to push the quadrotor to agile maneuvers. 
Similarly, Paparazzi~\cite{paparazzi} is an open-source project for drones, which supports various hardware platforms.
However, the supported autopilots have very limited onboard compute capability, rendering them unsuited for agile autonomous flight.
The Crazyflie~\cite{GiernackiCrazy} is an extremely lightweight quadrotor platform with a takeoff weight of only \SI{27}{\gram}. 
The minimal hardware setup leaves no margin for additional sensing or computation, prohibiting any non-trivial navigation task.

\begin{figure*}[!ht]
    \centering
    \vspace*{-12pt}
    \includegraphics[width=1\linewidth]{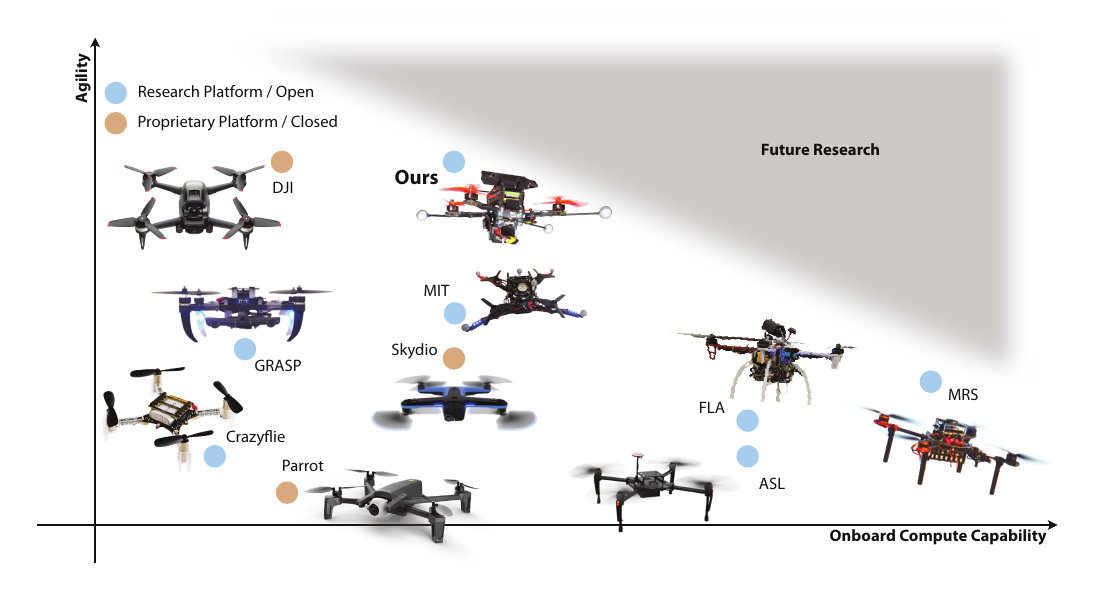}
    \vspace*{-18pt}
    
    \def\arraystretch{1.1}
	\resizebox{\textwidth}{!}{%
        \begin{tabular}{@{}l|c|c|c|c|r|c|r|r}
        	\toprule
        	framework & open-source & simulation & \parbox{2.0cm}{\centering onboard computer} & \parbox{2.0cm}{\centering low-level controller} & \parbox{3.0cm}{\centering CPU mark \\ (higher is better)} & GPU & \parbox{3.15cm}{maximum speed \\ (\SI{0}{}-\SI{100}{\km\per\hour} time)} & thrust/weight\\
        	\midrule
        	PX4~\cite{meier2015px4} & SW and HW  & \textcolor{green}{\cmark} &  \textcolor{red}{\xmark}  & custom open source & - & \textcolor{red}{\xmark} & - & - \\
        	Paparazzi~\cite{paparazzi} & SW and HW & \textcolor{green}{\cmark} &  \textcolor{red}{\xmark} & custom open source & - & \textcolor{red}{\xmark} & - & - \\ 
        	\midrule[0.005em]
        	DJI~\cite{DJIFPV} & - & \textcolor{red}{\xmark} & \textcolor{red}{\xmark}  & proprietary & - & \textcolor{red}{\xmark} & \SI{140}{\km\per\hour} ~~ (\SI{2.0}{\second}) & $\approx4.434$ \\
        	Skydio~\cite{Skydio} & -  & \textcolor{red}{\xmark} & \textcolor{red}{\xmark}  & proprietary &  - & \textcolor{red}{\xmark} & \SI{58}{\km\per\hour}  & -  \\
        	Parrot~\cite{ParrotANAFI_AI} & SW  & \textcolor{red}{\xmark}  & \textcolor{red}{\xmark} & proprietary & - & \textcolor{red}{\xmark} & \SI{55}{\km\per\hour} & -  \\
        	Crazyflie~\cite{GiernackiCrazy} & SW and HW & \textcolor{green}{\cmark} & \textcolor{red}{\xmark} & custom open source & - & \textcolor{red}{\xmark} & - & $\approx2.26$  \\
        	FLA-Quad~\cite{Mohta18jfr} & SW and HW & \textcolor{red}{\xmark} & \textcolor{green}{\cmark} & PX4 & $3{,}383$ & \textcolor{red}{\xmark} & - & $\approx2.38$  \\
        	GRASP-Quad~\cite{loianno2016estimation} & - & \textcolor{red}{\xmark} & \textcolor{green}{\cmark} & custom & $625$ & \textcolor{red}{\xmark} & - & $\approx1.80$  \\
        	MIT-Quad~\cite{antonini202IJRR}  & - & \textcolor{red}{\xmark} & \textcolor{green}{\cmark} & custom & $1{,}343$ & \textcolor{green}{\cmark} & - & $\approx2.33$  \\
        	ASL-Flight~\cite{Sa2018ram} & SW and HW & \textcolor{red}{\xmark} & \textcolor{green}{\cmark} & DJI & $3{,}383$ & \textcolor{red}{\xmark} & - & $\approx2.32$  \\
        	RPG-Quad~\cite{Faessler18ral} & SW and HW & \textcolor{green}{\cmark} & \textcolor{green}{\cmark}  & Betaflight & $633$ & \textcolor{red}{\xmark} & - & $\approx4.00$  \\
        	MRS UAV~\cite{Baca2021jirs} & SW and HW & \textcolor{green}{\cmark} & \textcolor{green}{\cmark} & PX4 & $8{,}846$ & \textcolor{red}{\xmark} & - & $\approx2.50$  \\
        	\midrule
        	\textbf{Agilicious (Ours)} & SW and HW & \textcolor{green}{\cmark} & \textcolor{green}{\cmark} & custom open source & $1{,}343$ & \textcolor{green}{\cmark} & \SI{131}{\km\per\hour} ~~ (\SI{1.01}{\second}) & $\approx5.00$  \\
        	\bottomrule
        \end{tabular}
        }
    \caption{
    A comparison of different available consumer and research platforms with respect to available onboard compute capability and agility. 
    The platforms are compared based on their openness to the community, support of simulation and onboard computation, used low-level controller, CPU~power (reported according to publicly available benchmarks {\small\url{https://www.cpubenchmark.net}} and corresponding to the speed of solving a set of benchmark algorithms that represent a generic program), and the availability of onboard general-purpose GPU.
	The agility of the platforms is expressed in the terms of \acl{TWR}; however, we also report the maximal velocity as an agility indicator due to limited information about the commercial platforms. 
	\rebuttal{The PX4~\cite{meier2015px4} and the Paparazzi~\cite{paparazzi} are rather low-level autopilot frameworks without high-level computation capability, but they can be integrated in other high-level frameworks~\cite{Mohta18jfr,Baca2021jirs}.}
    The open-source frameworks FLA~\cite{Mohta18jfr}, ASL\cite{Sa2018ram}, and MRS~\cite{Baca2021jirs} have relatively large weight and low agility.
    The DJI~\cite{DJIFPV}, Skydio~\cite{Skydio}, and Parrot~\cite{ParrotANAFI_AI} are closed-source commercial products that are not intended for research purposes.
    The Crazyflie~\cite{GiernackiCrazy} does not allow for sufficient onboard compute or sensing, while the MIT~\cite{antonini202IJRR} and GRASP~\cite{loianno2016estimation} platforms are not available open-source.
    Finally, our proposed Agilicious framework provides agile flight performance, onboard GPU-accelerated compute capabilities, as well as open-source and open-hardware availability. 
    }
    \label{fig:comparison}
    \vspace*{-6pt}
\end{figure*}

To address the requirements of agile flight, the shortcomings of existing works, and to enable the research community to progress fast towards agile flight, we present an open-source and open-hardware quadrotor platform for agile flight at more than 5\si{\g} acceleration while providing substantial onboard compute and versatile software.
The hardware design leverages recent advances in motor, battery, and frame design initiated by the \ac{FPV} racing community. 
The design objectives resulted in creating a lightweight \SI{750}{\gram} platform with maximal speed of \SI{131}{\km\per\hour}.
This high-performance drone hardware is combined with a powerful onboard computer that features an integrated GPU, enabling complex neural network architectures to run at high frequency while concurrently running optimization-based control algorithms at low latency onboard the drone.
The most important features of the Agilicious framework are summarized and compared with relevant research and industrial platforms in Figure~\ref{fig:comparison}.
A qualitative comparison of mutually contradicting onboard computational power and agility is presented in Figure~\ref{fig:comparison}.

\pagebreak

\newpage
In co-design with the hardware, we complete the drone design with a modular and versatile software stack, called Agilicious.
It provides a software architecture that allows to easily transfer algorithms from prototyping in simulation to real-world deployment in instrumented experiment setups, and even pure onboard-sensing applications in unknown and unstructured environments. 

This modularity is key for fast development and high-quality research, since it allows to quickly substitute existing components with new prototypes and enables all software components to be used in standalone testing applications, experiments, benchmarks, or even completely different applications. 

The hardware and software presented in this work have been developed, tested, and refined throughout a series of research publications~\cite{falanga2018pampc,kaufmann18icra,kaufmann2020RSS,loquercio2021wild,sun2021autonomous,Foehn2021science,Nisar2019ral,song2021autonomous}.
All these publications share the ambition to push autonomous drones to their physical limits. 
The experiments, performed in a diverse set of environments demonstrate the versatility of Agilicious by deploying different estimation, control, and planning strategies on a single physical platform. 
The flexibility to easily combine and replace both hard- and software components in the flight stack while operating on a standardized platform facilitates testing new algorithms and accelerates research on fast autonomous flight.


\section{Results}  \label{sec:experiments}
\rebuttal{
Our experiments, conducted in simulation and in the real world, demonstrate that Agilicious can be used to perform cutting-edge research in the fields of agile quadrotor control, quadrotor trajectory planning, and learning-based quadrotor research. 
We evaluate the capabilities of the Agilicious software and hardware stack in a large set of experiments that investigate trajectory tracking performance, latency of the pipeline, combinations of Agilicious with a set of commercially or openly available vision-based state estimators.
Finally, we present two demonstrators of recent research projects that build on Agilicious. 
}
\vspace*{-6pt}
\subsection{Trajectory Tracking Performance}
\label{tracking_arena_flight}

\begin{figure*}[t!]
    \centering
    \includegraphics[width=\textwidth]{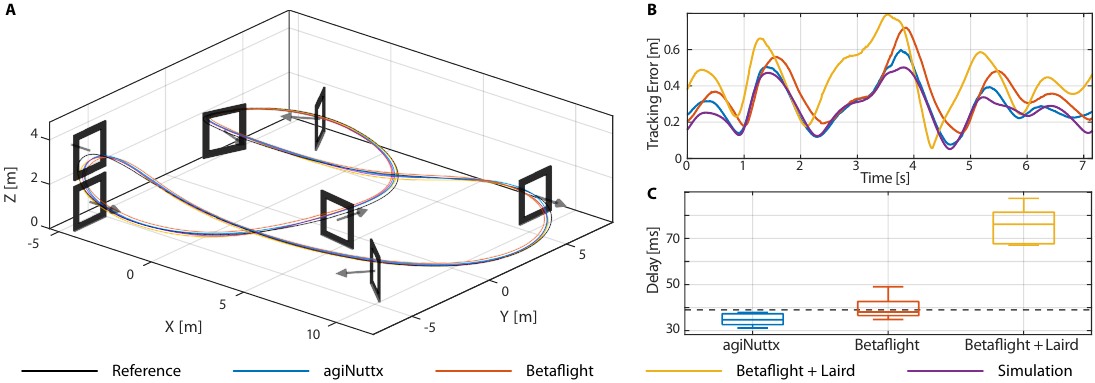}
    \caption{
    \rebuttal{
    An agile trajectory with speeds up to 60\si{\kilo\meter\per\hour} and an acceleration of 4\si{\g}, executed in an indoor instrumented flight volume.
    We compare multiple different drone configurations, including our own low-level flight controller software \agilow{}, an off-the-shelf BetaFlight controller, the BetaFlight controller together with offboard computing and remote control through a Laird wireless transmitter, and our included simulation.
    Figure A depicts an overview of the flown trajectory.
    Figure B shows the tracking errors along a single lap over all three configurations.
    Our provided \agilow{} achieves the best tracking performance, followed by BetaFlight combined with onboard computation.
    In contrast, offloading computation from the drone and controlling it remotely significantly impacts performance.
    This is due to the massively increased latency, depicted in Figure C, where, for reference, the motor time constant of 39.1\si{\milli\second} is marked as a dashed line ({\color{gray}-\,-\,-}).
    Additionally, in Figure B it is visible that the simulation exhibits very similar error characteristics, thanks to our accurate aerodynamic modelling.
    }
    }
    \label{fig:tracking}
    \vspace*{-12pt}
\end{figure*}

\rebuttal{
In this section, we demonstrate the tracking performance of our platform by flying an aggressive time-optimal trajectory in a drone racing scenario.
Additionally, to benchmark our planning and control algorithms, we compete against a world-class drone racing pilot \ac{FPV} pilot, reported in~\cite{Foehn2021science}.
As illustrated in Figure~\ref{fig:tracking}, our drone racing track consists of seven gates that need to be traversed in a pre-defined order as fast as possible.
The trajectory used for this evaluation reaches speeds of 60\si{\kilo\meter\per\hour} and accelerations of 4~\si{\g}.

Flying through gates at such high speed requires precise state estimates, which is still an open challenge using vision-based state estimators~\cite{delmerico19icra}.
For this reason, we conduct these experiments in an instrumented flight volume of $30\times30~\times$~\SI{8}{\meter} (\SI{7200}{\cubic\meter}), equipped with 36 VICON cameras that provide precise pose measurements at \SI{400}{\hertz}.
%
However, even when provided with precise state estimation, accurately tracking such aggressive trajectories poses considerable challenges with respect to the controller design, which usually requires several iterations of algorithm development and substantial tuning effort.
The proposed Agilicious flight stack allows us to easily design, test, and deploy different control methods by first verifying them in simulation and then fine-tuning them in the real-world. 
The transition from simulation to real-world deployment requires no source-code changes or adaptions, which reduces the risk of crashing expensive hardware, and is one of the major features of Agilicious accelerating rapid-prototyping.
Figure~\ref{fig:tracking} includes a simulated flight that shows similar characteristics and error statistics compared to the real-world flights described next.

We evaluate three different system and control approaches including onboard computation with an off-the-shelf BetaFlight~\cite{Betaflight} flight controller, our custom open-source \agilow{} controller, and an offboard-control scenario.
These three system configurations represent various use cases of Agilicious, such as running state-of-the-art single-rotor control onboard the drone using our \agilow{} described in Sec.~\ref{sec:flight_hardware}, or simple remote control by executing Agilicious on a desktop computer and forwarding the commands to the drone.
All configurations use the motion capture state estimate and our single-rotor \ac{MPC} described in Section~\ref{sec:controller}~\cite{Sun2021arxiv} as high-level controller.
We use the single-rotor thrust formulation to correctly account for actuation limits, but use bodyrates and collective thrust as command modality.

The first configuration runs completely onboard the drone with an additional low-level controller in the form of \ac{INDI} as described in Section~\ref{sec:controller}~\cite{Sun2021arxiv}.
It uses the MPC's output to compute refined single-rotor thrust commands using \ac{INDI}, to reduce the sensitivity to model inaccuracies.
These single-rotor commands are executed using our \agilow{} flight controller with closed-loop motor speed tracking.

The second configuration also runs onboard and directly forwards the bodyrate and collective thrust command from the \ac{MPC} to a BetaFlight~\cite{Betaflight} controller.
This represents the most simplistic system which does not require flashing the flight controller and is compatible with a wide range of readily available off-the-shelf hobbyist drone components.
However, in this configuration, the user does not get any \ac{IMU} or motor speed feedback, as those are not streamed by BetaFlight.

The third configuration is equal to the second configuration with the difference that the Agilicious flight stack runs offboard on a desktop or laptop computer.
The bodyrate and collective thrust commands from the \ac{MPC} are streamed to the drone using a serial wireless link implemented through LAIRD~\cite{LAIRD}.
This configuration allows to run computationally demanding algorithms, such as GPU-accelerated neural networks, with minimal modifications.
However, due to the additional wireless command transmission, there is a higher latency which can potentially degrade the control performance.

Finally, the Agilicious simulation is executed using the same setup as the first configuration.
It uses accurate models for the quadrotor and motor dynamics, as well as a \acl{BEM} aerodynamic model as described in \cite{bauersfeld2021neurobem}.

\begin{figure*}[t!]
    \centering
    \includegraphics[width=\linewidth]{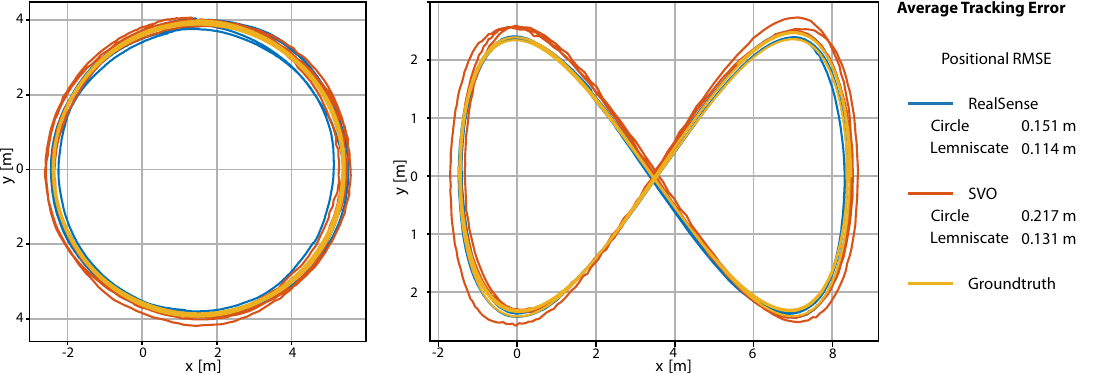}
    \caption{\rebuttal{A comparison of two different \acl{VIO} solutions.
    The first solution consists of the Intel RealSense T265, an off-the-shelf sensor featuring a stereo camera, an \ac{IMU}, and an integrated \ac{VIO} pipeline running on the integrated compute hardware.
    The second solution consists of a monocular camera, the \ac{IMU} of the onboard flight controller, and SVO \cite{Forster17troSVO}.
    The estimates of both solutions are compared against motion-capture ground truth using~\cite{Zhang18iros} on a circle and a Lemniscate trajectory, flown using Agilicious in an indoor environment.
    Both systems show accurate tracking performance and could be used as cost-effective drop-in replacement for motion capture systems and enable deployment in the wild.
    While the Intel RealSense T265 is a convenient off-the-shelf option, using other cameras in combination with the onboard flight controller \ac{IMU} and an open-source or custom \ac{VIO} pipeline enables tailored solutions and research-oriented data access.}}
    \label{fig:vio}
    \vspace*{-12pt}
\end{figure*}

Figure~\ref{fig:tracking}A,B depict the results of these trajectory tracking experiments.
Our first proposed configuration (i.e. with onboard computation and the custom \agilow{} flight controller) achieves the best overall tracking performance with the lowest average positional \ac{RMSE} of just 0.322\si{\meter} at up to 60~\si{\kilo\meter\per\hour} and 4~\si{\g}.
Next up is the second configuration with BetaFlight, still achieving less than 0.385\si{\meter} average positional \ac{RMSE}.
Finally, the third configuration with offboard control exhibits higher latency, leading to an increased average positional \ac{RMSE} of 0.474\si{\meter}.
As can be seen, our simulation closely matches the performance observed in real world in the first configuration.
The simulated tracking results in slightly lower errors, 0.320\si{\meter} \ac{RMSE}, since even the state-of-the-art aerodynamic models \cite{bauersfeld2021neurobem} fail to reproduce the highly non-linear and chaotic real aerodynamics.
This simulation accuracy allows a seamless transition from simulation prototyping to real-world verification, and is one of the prominent advantages of Agilicious.

Additional experiments motivating the choice of \ac{MPC} as outer-loop controller and its combination with \ac{INDI} can be found in~\cite{Sun2021arxiv}, details on the planning of the time-optimal reference trajectory are elaborated on in \cite{Foehn2021science}, and additional extensions to the provided \ac{MPC} for fast flight are in \cite{Romero2021arxiv} and for rotor-failure \ac{MPC} in \cite{Nan21arxiv}.
These related publications also showcase performance at even higher speeds of up to 70~\si{\kilo\meter\per\hour} and accelerations reaching 5~\si{\g}.
The following section gives further insights into the latency of the three configurations tested herein, including on- and offboard control architecture, as well as BetaFlight and \agilow{} flight controllers.
}

\begin{figure*}[t!]
    \centering
    \includegraphics[width=1\linewidth]{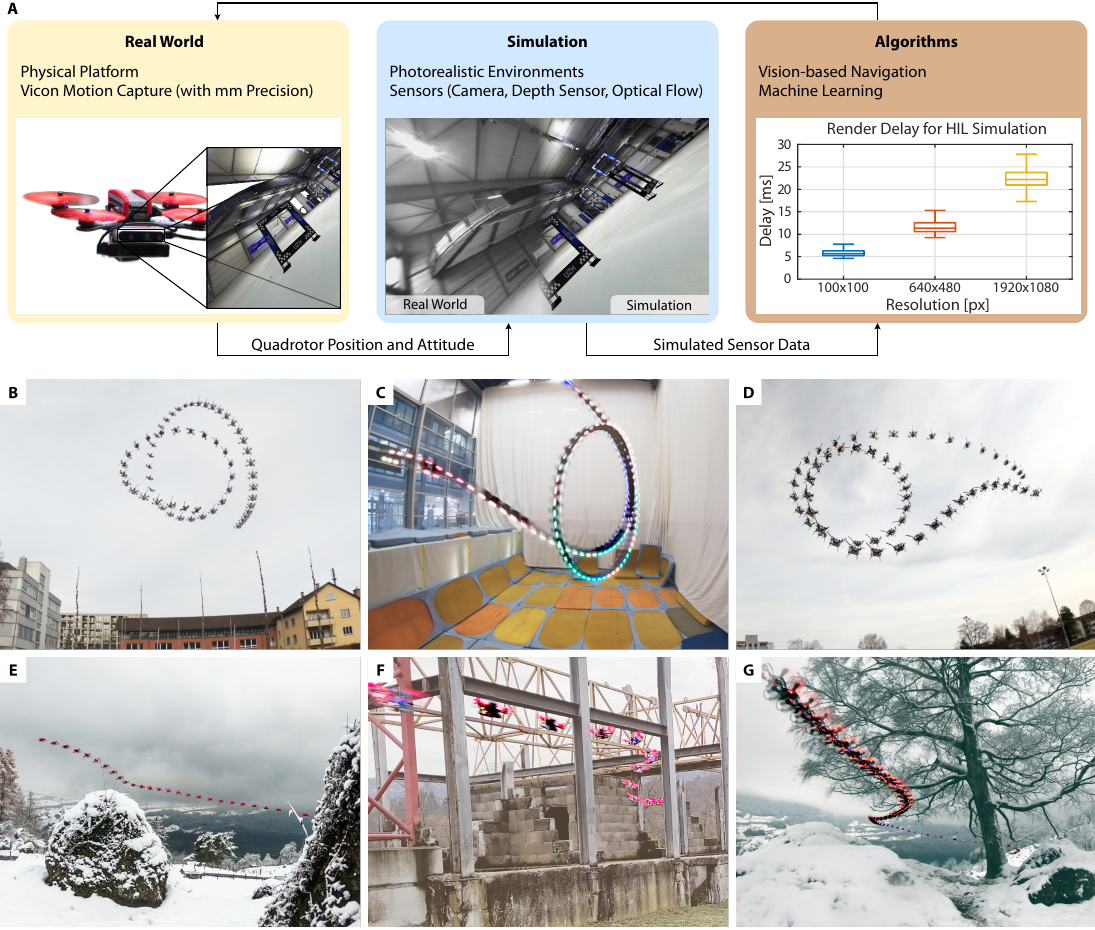}
    \caption{
    \rebuttal{
    A: The hardware-in-the-loop simulation of Agilicious consists of a real quadrotor flying in a motion capture system combined with photorealistic simulation of complex 3D environments. 
    Multiple sensors can be simulated with minimal delays while virtually flying in various simulated scenes. 
    Such hardware-in-the-loop simulation offers a modular framework for prototyping robust vision-based algorithms safely, efficiently, and inexpensively.
    B-G: The Agilicious platform is deployed in a diverse set of environments while only relying on onboard sensing and computation. 
    B-D: The quadrotor performs a set of acrobatic maneuvers using a learned control policy. 
    E-G: By leveraging zero-shot sim-to-real transfer, the quadrotor platform performs agile navigation through cluttered environments. Courtesy of Kaufmann et al.~\cite{kaufmann2020RSS} and Loquercio et al.~\cite{loquercio2021wild}
    }
    }
    \label{fig:demonstrators}
    \vspace*{-12pt}
\end{figure*}

\subsection{Control Latency}
\rebuttal{
All real systems with finite resources suffer from communication and computation delays, while dynamic systems and even filters can introduce additional latency and bandwidth limitations.
Analyzing and minimizing these delays is fundamental for the performance in any control task, especially when tracking agile and fast trajectories in the presence of model mismatch, disturbances and actuator limitations.
In this section we conduct a series of experiments that aim to analyse and determine the control latency, from command to actuation, of the proposed architecture for the three different choices of low level configurations: our \agilow{}, BetaFlight, and BetaFlight with offboard control.

For this experiment, the quadrotor has been mounted on a load cell (ATI Mini 40 SI-20-1~\cite{LoadCell}) measuring the force and moment acting on the platform.
To measure the latency, a collective thrust step command of 12\si{\newton} is sent to the corresponding low level controller, while measuring the exerted force on the drone.
These force sensor measurements are time-synchronized with the collective thrust commands, and fitted through a first-order system representing to motor dynamics.
The measured delays are reported in Figure~\ref{fig:tracking}C as the difference between the time at which the high level controller sends the collective thrust command, and the time at which the measured force effectively starts changing.
The results show that both \agilow{} has the lowest latency at 35\si{\milli\second}, with BetaFlight slightly slower at 40.15 \si{\milli\second}.
A large delay can be observed when using offboard control and sending the commands via Laird connection to the drone, in which case the latency rises to more than 75\si{\milli\second}.
The impact of these latencies is also reflected in the tracking error in Figure~\ref{fig:tracking}A,B.
To put the measured delays into perspective, the motor's time constant of 39.1\si{\milli\second}, which dictates the actuator bandwidth limitations, is indicated in Figure~\ref{fig:tracking}C.
Finally, Section~\ref{sec:hil} gives some insight into the latencies introduced when using Agilicious together with Flightmare~\cite{yunlong2020flightmare} in a hardware-in-the-loop setup.
}

\subsection{Visual-Inertial State Estimation}

\rebuttal{
Deploying agile quadrotors outside of instrumented environments requires access to onboard state estimation.
There exist many different approaches including GPS, lidar, and vision-based solutions.
However, for size and weight constrained aerial vehicles, visual-inertial odometry has proven to be the go-to solution because of the sensors' complementary measurement modalities, low cost, mechanical simplicity, and reusability for other purposes, such as depth estimation for obstacle avoidance.

The Agilicious platform provides a versatile sensor mount that is compatible with different sensors and can be easily adapted to fit custom sensor setups. 
In this work, two different \ac{VIO} solutions are evaluated: (i)~the proprietary, off-the-shelf Intel RealSense T265 and (ii)~a simple camera together with the onboard flight controller IMU and an open-source \ac{VIO} pipeline in the form of SVO Pro~\cite{Forster17troSVO,SVO} with its sliding-window backend.
While the RealSense T265 performs all computation on-chip and directly provides a state estimate via USB3.0, the alternative VIO solution uses the Jetson TX2 to run the VIO software and allows researchers to interface and modify the state estimation software.
Specifically, for sensor setup~(ii), a single Sony IMX-287 camera at 30\si{\hertz} with a 165\textdegree\, diagonal field-of-view is used, combined with the \ac{IMU} measurements of the flight controller at 500\si{\hertz}, calibrated using the Kalibr toolbox~\cite{Rehder2016icra}.

To verify their usability, a direct comparison of both \ac{VIO} solutions with respect to ground-truth is provided. 
Performance is evaluated based on the estimation error~\cite{Zhang18iros} obtained on two trajectories flown with Agilicious.
The flown trajectories consist of a circle trajectory with radius of 4\si{\meter} at a speed of 5\si{\meter\per\second}, and a Lemniscate trajectory with an amplitude of 5\si{\meter} at a speed of up to 7\si{\meter\per\second}.

Figure~\ref{fig:vio} shows the performance of both \ac{VIO} solutions in an $xy$-overview of the trajectories together with their \acf{ATE} \ac{RMSE}.
Both approaches perform well on both trajectories, with the Intel RealSense achieving slightly better accuracy according to the \ac{ATE} of 0.151\si{\meter} on the circle and 0.114\si{\meter} on the Lemniscate, compared to the monocular SVO approach with 0.217\si{\meter} and 0.131\si{\meter}, respectively.
This is expected since the Intel RealSense uses a stereo camera plus \ac{IMU} setup and is a fully integrated solution, while sensor setup (ii) aims at minimal cost by only adding a single camera and otherwise exploiting the existing flight-controller IMU and onboard compute resources.

However, at the timing of writing this manuscript, the Intel RealSense T265 is being discontinued.
Other possible solutions include camera sensors such as
SevenSense~\cite{SevenSense},
MYNT EYE~\cite{MyntEye},
or MatrixVision~\cite{BlueFox},
and other stand-alone cameras,
combined with software frameworks like
ArtiSense~\cite{ArtiSense},
SlamCore~\cite{SlamCore},
or open-source frameworks like
VINSmono~\cite{Qin17tro},
OpenVINS~\cite{Geneva2020icra},
or SVO Pro~\cite{Forster17troSVO,SVO}, evaluated in \cite{Delmerico2018icra}.
Furthermore, there are other fully integrated alternatives to the Intel RealSense~\cite{Realsense_data_sheet}, including the
Roboception~\cite{Roboception} and the ModalAI Voxl CAM~\cite{ModalAI}.
}

\subsection{Demonstrators}
\rebuttal{
The Agilicious software and hardware stack is intended as a flexible research platform. 
To illustrate its broad applicability, this section showcases a set of research projects that have been enabled through Agilicious. 
Specifically, we demonstrate the performance of our platform in two different experimental setups covering \acl{HIL} simulation and autonomous flight in the wild using only onboard sensing. 
}

\subsubsection{Hardware in the Loop Simulation}
\label{sec:hil}

\rebuttal{
Developing vision-based perception and navigation algorithms for agile flight is not only slow and time-consuming, due to the large amount of data required to train and test perception algorithms in diverse settings, but it progressively becomes less safe and more expensive as more aggressive flights can lead to devastating crashes.
This motivates the Agilicious framework to support hardware-in-the-loop simulation, which consists of flying a physical quadrotor in a motion-capture system while observing virtual photorealistic environments, as previously shown in~\cite{guerra2019flightgoggles}.
The key advantage of hardware-in-the-loop simulation over classical synthetic experiments~\cite{Furrer2016rotors} is the usage of real-world dynamics and proprioceptive sensors, instead of idealized virtual devices, combined with the ability to simulate arbitrarily sparse or dense environments without the risk of crashing into real obstacles.

The simulation of complex 3D environments and realistic exteroceptive sensors is achieved using our high-fidelity quadrotor simulator~\cite{yunlong2020flightmare} built on Unity~\cite{juliani2018unity}.
The simulator can offer a rich and configurable sensor suite, including RGB cameras, depth sensors, optical flow, and image segmentation, combined with variable sensor noise levels, motion blur, distortion and diverse environmental factors such as wind and rain.
The simulator achieves this by introducing only minimal delays (see Figure~\ref{fig:demonstrators}A), ranging from 13\si{\milli\second} for 640\texttimes480 VGA resolution to 22\si{\milli\second} for 1920\texttimes1080 full HD images, when rendered on a NVIDIA RTX 2080 GPU.

Overall, the integration of our agile quadrotor platform and high-fidelity visual simulation provides an efficient framework for the rapid development of vision-based navigation systems in complex and unstructured environments.
}

\subsubsection{Vision-based Agile Flight with Onboard Sensing and Compute}

\rebuttal{
When a quadrotor can only rely on onboard vision and computation, perception needs to be effective, low-latency, and robust to disturbances.
Violating this requirement may lead to crashes, especially during agile and fast maneuvers where latency has to be low and robustness to perception disturbances and noise must be high.
However, vision systems either exhibit reduced accuracy or completely fail at high speeds due to motion blur, large pixel displacements, and quick illumination changes~\cite{cadena2016past}.
To overcome these challenges, vision-based navigation systems generally build upon two different paradigms.
The first uses the traditional perception-planning-and-control pipeline, represented by standalone blocks which are executed in sequence and designed independently~\cite{Oleynikova17IROS, florence2020integrated, Falanga17ICRA, zhou2019robust, FalMueFaeSca17, loianno2016estimation}.
Works in the second category substitute either parts or the complete perception-planning-and-control pipeline with learning-based methods~\cite{zeng2019end, zeng2020dsdnet, Bansal19Corl, homayounfar2019cvpr, kaufmann2020RSS, Zhang18icra, Ross13icra, SadeghiL17Rss, Loquercio18ral, Gandhi17Iros}.

The Agilicious flight stack supports both paradigms and has been used to compare traditional and learning-based methods on agile navigation tasks in unstructured and structured environments (see Figure~\ref{fig:demonstrators}).
Specifically, Agilicious facilitated quantitative analyses of approaches for autonomous acrobatic maneuvers~\cite{kaufmann2020RSS}(Fig.~\ref{fig:demonstrators}B-D) and high-speed obstacle avoidance in previously unknown environments~\cite{loquercio2021wild}(Fig.~\ref{fig:demonstrators}E-G). 
Both comparisons feature a rich set of approaches consisting of traditional planning and control~\cite{falanga2018pampc, florence2020integrated, zhou2019robust} as well as learning-based methods~\cite{kaufmann2020RSS, loquercio2021wild} with different input and output modalities.
Thanks to its flexibility, Agilicious enables an objective comparison of these approaches on a unified control stack, without biasing results due to different low-level control strategies.

}


\section{Discussion} \label{sec:discussion}

The presented Agilicious framework substantially advances the published state of the art in autonomous quadrotor platform research. 
It offers advanced computing capabilities combined with the most powerful open-source and open-hardware quadrotor platform created to date, opening the door for research on the next generation of autonomous robots.
We see three main axes for future research based on our work.

First, we hypothesize that future flying robots will be smaller, lighter, cheaper, and consuming less power than what is possible today, increasing battery life, crash-resilience, as well as \acl{TWR} and \acl{TIR}~\cite{Floreano2015nature}.
This miniaturization is evident in state-of-the-art research towards direct hardware implementations of modern algorithms in the form of \ac{ASIC}s, such as the Navion~\cite{Suleiman2019jssc}, the Movidius \cite{Movidius}, or the PULP processor~\cite{Palossi18arxiv, Palossi2019dcoss}.
These highly specialized in-silicon implementations are typically magnitudes smaller and more efficient than general compute units.
Their success is rooted in the specific structure many algorithmic problems exhibit, such as the parallel nature of image data or the factor-graph representations used in estimation, planning, and control algorithms, like SLAM, \acl{MPC}, and neural network inference.

Second, the presented framework was mainly demonstrated with fixed-shape quadrotors.
This is an advantage as the platform is easier to model and control, and less susceptible to hardware failures.
Nevertheless, platforms with a dynamic morphology are by design more adaptable to the environment and potentially more power efficient~\cite{Ajanic20sr, Chang20sr, Falanga18foldable, Bauersfeld20VTOL}.
For example, to increase flight time, a quadrotor might transform to a fixed-wing aircraft~\cite{Dsa16icra}.
Due to its flexibility, Agilicious is the ideal tool for the future development of morphable and soft aerial systems.

Finally, vision-based agile flight is still in the early stages and has not yet reached the performance of professional human pilots.
The main challenges lie in handling complex aerodynamics, e.g. transient torques or rotor inflow, low-latency perception and state estimation, and recovery from failures at high speeds.
In the last few years, considerable progress has been made by leveraging data-driven algorithms~\cite{bauersfeld2021neurobem, torrente2021ral, kaufmann2020RSS, loquercio2021wild} and novel sensors as event-based cameras~\cite{Falanga20SR, sun2021autonomous}, that provide a high dynamic range, low latency, and low battery consumption~\cite{Gallego20pami}.
A major opportunity for future work is to complement the existing capabilities of Agilicious with novel compute devices such as the Intel Loihi~\cite{Davies2018micro,decroonLoihi20Arxiv,vitale21icra} or SynSense Dynap~\cite{Moradi2018dynapse} neuromorphic processing architecture, which are specifically designed to operate in an event-driven compute scheme.
Due to the modular nature of Agilicious, individual software components can be replaced by these novel computing architectures, supporting rapid iteration and testing.

In summary, Agilicious offers a unique quadrotor testbed to accelerate current and future research on fast autonomous flight. 
Its versatility in both hardware and software allows deployment in a wide variety of tasks, ranging from exploration or search and rescue missions to acrobatic flight. 
Furthermore, the modularity of the hardware setup allows integrating novel sensors or even novel compute hardware, enabling to test such hardware on an autonomous agile vehicle.
By open-sourcing Agilicious, we provide the research and industrial community access to a highly agile, versatile, and extendable quadrotor platform.


\section{Materials and Methods}
\label{sec:materials}

Designing a versatile and agile quadrotor research platform requires to co-design its hardware and software, while carefully trading off competing design objectives such as onboard computing capacity and platform agility.
In the following, the design choices that resulted in the flight hardware, compute hardware, and software design of Agilicious (see Figure~\ref{fig:overview}) are explained in detail.

\subsection{Compute Hardware}\label{sec:compute_hardware}
To exploit the full potential of highly unstable quadrotor dynamics, a high-frequency low-latency controller is needed.
Both of these requirements are difficult to meet with general-purpose operating systems, which typically come without any real-time execution guarantees.
Therefore, we deploy a low-level controller with limited compute capabilities but reliable real-time performance, which stabilizes high-bandwidth dynamics, such as the motor speeds or the vehicle's bodyrate.
This allows complementing the system with a general-purpose high-level compute unit that can run Linux for versatile software deployment, with significantly relaxed real-time requirements.

\mypara{High-Level Compute Board}
The high-level of the system architecture provides all the necessary compute performance to run the full flight stack, including estimation, planning, optimization-based control, neural network inference, or other demanding experimental applications.
Therefore, the main goal is to provide general-purpose computing power, while complying with the strict size and weight limits.
We evaluate a multitude of different compute modules made from system-on-a-chip (SoC) solutions since they allow inherently small footprints.
An overview is shown in Tab.~\ref{tab:hardware}.
We exclude the evaluation of two popular contenders:
(a) the Intel NUC platform, since it neither provides any size and weight advantage over the Jetson Xavier AGX nor provides a general-purpose GPU;
and (b) the Raspberry Pi compute modules since they do not offer any compute advantages over the Odroid and UpBoard, and no size and weight advantage over the NanoPi product family.

As we target general flight applications, fast prototyping, and experimentation, it is important to support a wide variety of software, which is why we chose a Linux-based system. 
TensorFlow~\cite{TensorFlow} and PyTorch~\cite{PyTorch} are some of the most prominent frameworks with hardware-accelerated neural network inference.
Both of them support accelerated inference on the Nvidia CUDA general-purpose GPU architecture, which renders nVidia products favorable, as other products have no or poorly-supported accelerators.
Therefore, four valid options remain, listed in the second row of Tab.~\ref{tab:hardware}.
While the Jetson Xavier AGX is beyond our size and weight goals, the Jetson Nano provides no advantage over the Xavier NX, rendering both the Jetson TX2 and Xavier NX viable solutions.
Since these two CUDA-enabled compute modules require breakout boards to connect to peripherals, our first choice is the TX2 due to the better availability and diversity of such adapter boards and their smaller footprint.
For the breakout board we recommend the ConnectTech Quasar~\cite{Quasar}, providing multiple USB ports, Ethernet, serial ports, and other interfaces for sensors and cameras.

\begin{table*}[t]
    \centering
    \def\arraystretch{1.2}
    \footnotesize
    \begin{tabularx}{1\linewidth}{l|Y|Y|Y|Y}
        \multicolumn{5}{c}{\small{without General-Purpose GPU}} \\
        \toprule
        &
            \includegraphics[width=0.6\linewidth]{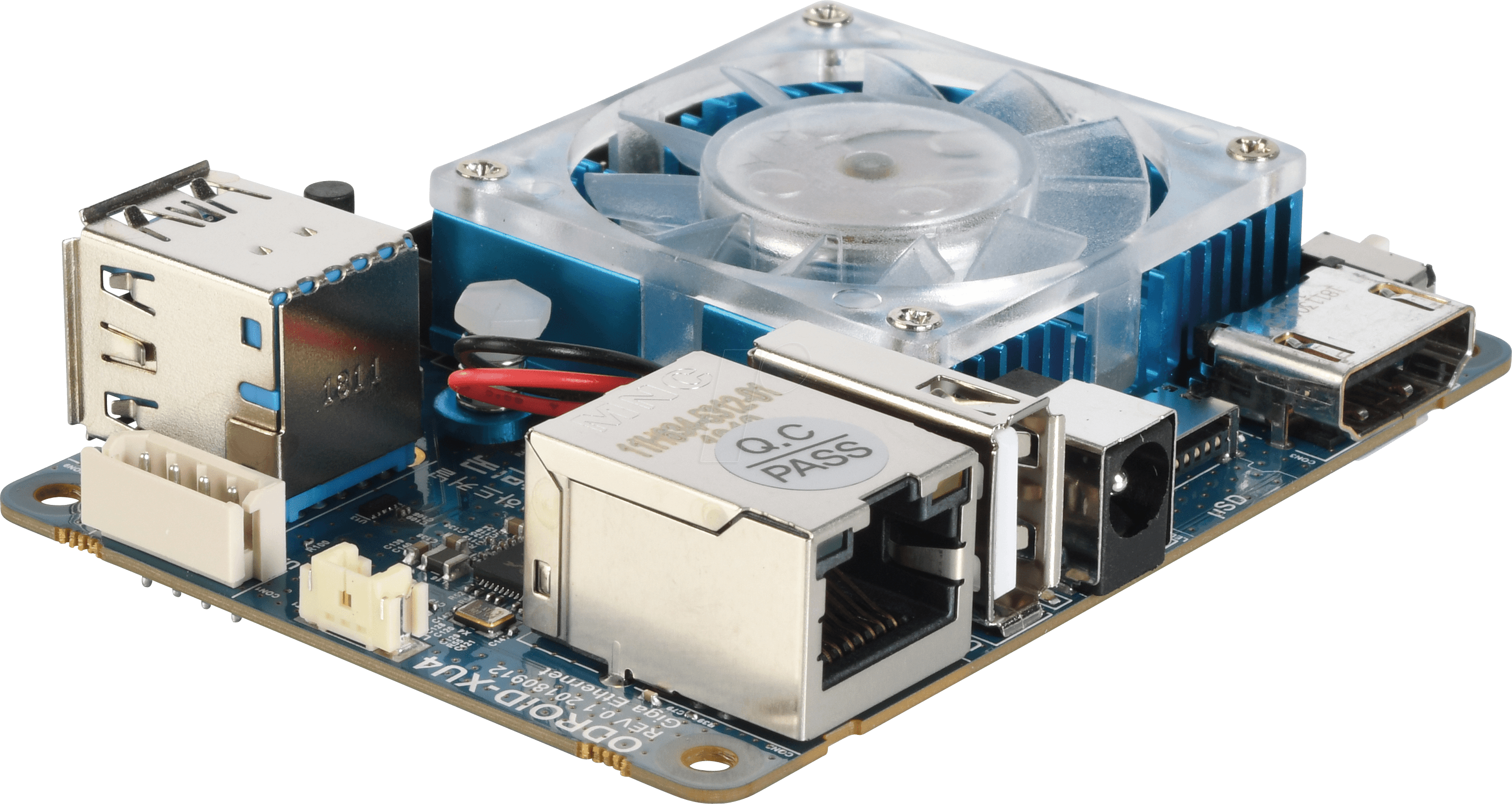} &
            \includegraphics[width=0.6\linewidth]{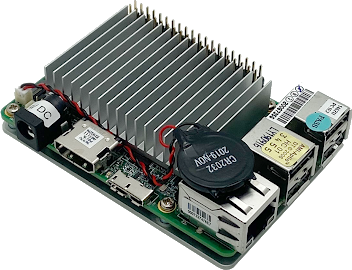} &
            \includegraphics[width=0.4\linewidth]{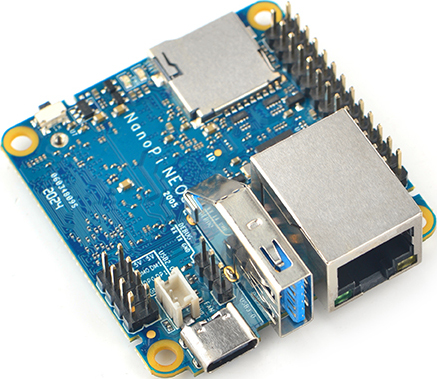} &
            \includegraphics[width=0.4\linewidth]{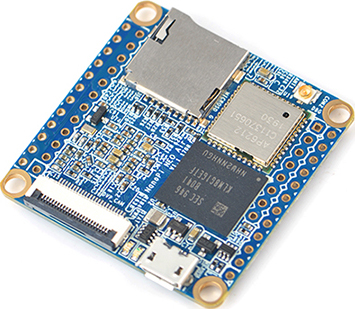} \\
        Product &
            \href{https://wiki.odroid.com/odroid-xu4/odroid-xu4}{Odroid XU4} &
            \href{https://up-board.org/up/specifications/}{Intel UpBoard} &
            \href{https://wiki.friendlyarm.com/wiki/index.php/NanoPi_NEO3}{NanoPi Neo 3} &
            \href{https://wiki.friendlyarm.com/wiki/index.php/NanoPi_NEO_Air}{NanoPi Neo air} \\
        \midrule
        CPU &
            8$\times$ 32-bit ARM \SI{2.1}{\giga\hertz} &
            4$\times$ 64-bit Atom \SI{1.92}{\giga\hertz} &
            4$\times$ 64-bit ARM \SI{1.5}{\giga\hertz} &
            4$\times$ 32-bit ARM \SI{1.2}{\giga\hertz} \\
        RAM &
            \SI{2}{\giga\byte} LPDDR3 &
            \SI{4}{\giga\byte} LPDDR3 &
            \SI{2}{\giga\byte} LPDDR4 &
            \SI{512}{\mega\byte} LPDDR3 \\
        GPU &
            Mali-T628 &
            Intel HD400 &
            Mali-450 MP4 &
            Mali-400 MP2 \\
        FLOPS &
            $\sim$\SI{120}{\giga\flops} &
            $\sim$ \SI{115}{\giga\flops} &
            $\sim$\SI{40}{\giga\flops} &
            $\sim$\SI{10}{\giga\flops} \\
        Storage &
            up to \SI{128}{\giga\byte} EMMC &
            up to \SI{64}{\giga\byte} EMMC &
            only SD card &
            \SI{8}{\giga\byte} EMMC \\
        Interfaces &
            USB, Ethernet, Serial, I2C, SPI, GPIO &
            USB, Ethernet, Serial, I2C, SPI, GPIO, 1 camera &
            USB, Ethernet, Serial, I2C, SPI, GPIO &
            USB, Ethernet, WIFI, Serial, I2C, SPI, GPIO, 1 camera \\
        Size &
            $83 \times 58 \times \SI{19}{\milli\meter}$ &
            $85 \times 57 \times \SI{20}{\milli\meter}$ &
            $40 \times 40 \times \SI{23}{\milli\meter}$ &
            $40 \times 40 \times \SI{10}{\milli\meter}$ \\
        Weight &
            \SI{59}{\gram} &
            \SI{79}{\gram} &
            \SI{36}{\gram} &
            \SI{24}{\gram} \\
        \bottomrule
        \multicolumn{5}{c}{}\\
        \multicolumn{5}{c}{\small{with General-Purpose GPU}} \\
        \toprule
        &
            \includegraphics[width=0.4\linewidth]{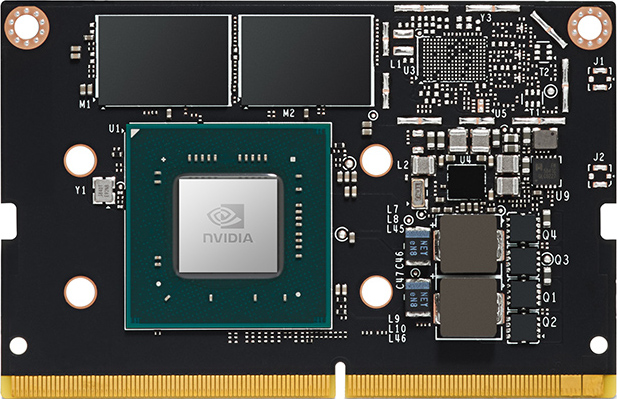} &
            \includegraphics[width=0.6\linewidth]{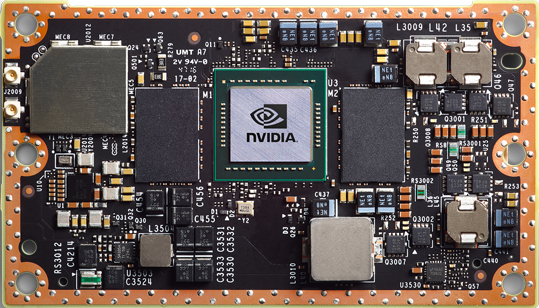} &
            \includegraphics[width=0.4\linewidth]{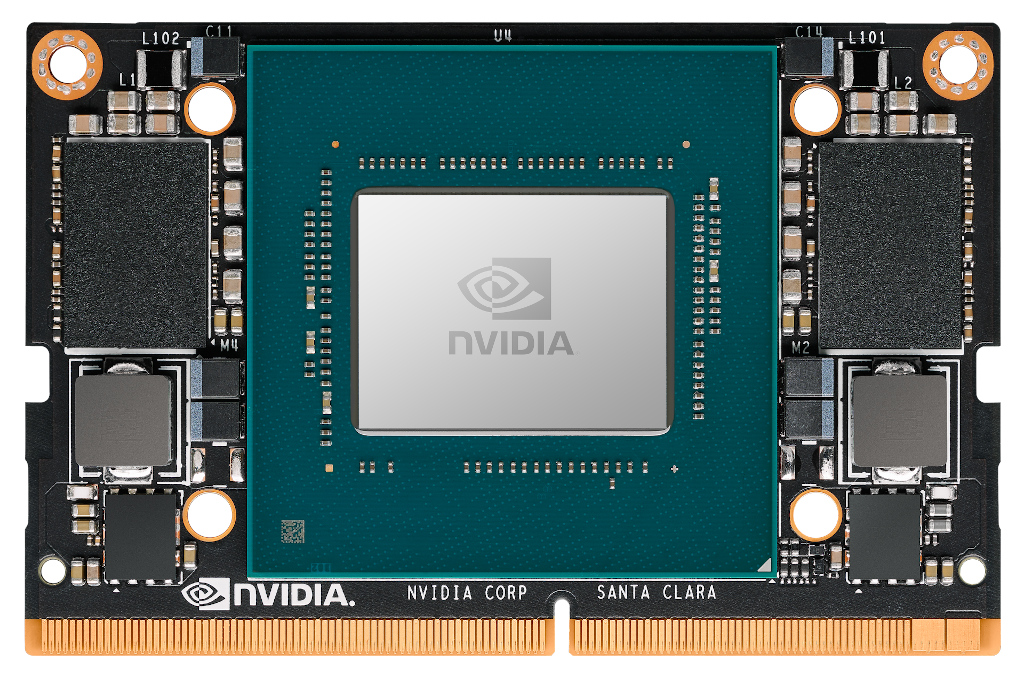} &
            \includegraphics[width=0.6\linewidth]{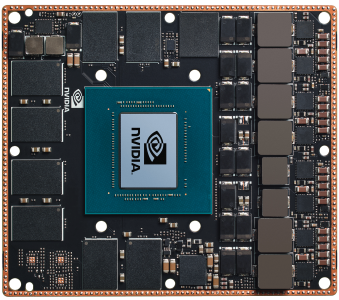} \\
        Product &
            \href{https://developer.nvidia.com/embedded/jetson_nano}{nVidia Jetson Nano} &
            \href{https://developer.nvidia.com/embedded/jetson_tx2}{nVidia Jetson TX2} &
            \href{https://developer.nvidia.com/embedded/jetson_xavier_nx}{nVidia Jetson Xavier NX} &
            \href{https://developer.nvidia.com/embedded/jetson-agx-xavier}{nVidia Jetson AGX Xavier} \\
        \midrule
        CPU &
            4$\times$ 64-bit ARM \SI{1.43}{\giga\hertz} &
            6$\times$ 64-bit ARM \SI{2.0}{\giga\hertz} &
            6$\times$ 64-bit ARM \SI{1.9}{\giga\hertz} &
            8$\times$ 64-bit ARM \SI{2.26}{\giga\hertz} \\
        RAM &
            4 \si{\giga\byte} LPDDR4 &
            8 \si{\giga\byte} LPDDR4 &
            8 \si{\giga\byte} LPDDR4 &
            32 \si{\giga\byte} LPDDR4 \\
        GPU &
            128$\times$ Maxwell CUDA &
            256$\times$ Pascal CUDA &
            384$\times$ Volta CUDA &
            512$\times$ Volta CUDA \\
        FLOPS &
            \SI{472}{\giga\flops} &
            \SI{1.33}{\tera\flops} &
            \SI{2.12}{\tera\flops} &
            \SI{11}{\tera\flops} \\
        Storage &
            \SI{16}{\giga\byte} EMMC &
            \SI{32}{\giga\byte} EMMC &
            \SI{16}{\giga\byte} EMMC &
            \SI{32}{\giga\byte} EMMC \\
        Interfaces &
            USB, Ethernet, Serial, I2C, SPI, GPIO, 4 cameras &
            USB, Ethernet, WIFI, Serial, I2C, SPI, GPIO, 6 cameras &
            USB, Ethernet, Serial, I2C, SPI, GPIO, 6 cameras &
            USB, Ethernet, Serial, I2C, SPI, GPIO, 6 cameras \\
        Size &
            $69.9 \times 45 \times \SI{22}{\milli\meter}$ &
            $87 \times 50 \times \SI{34}{\milli\meter}$ &
            $69.9 \times 45 \times \SI{22}{\milli\meter}$ &
            $100 \times 87 \times \SI{58}{\milli\meter}$ \\
        Weight &
            \SI{63}{\gram} &
            \SI{154}{\gram} &
            \SI{79}{\gram} &
            \SI{650}{\gram} \\
        \bottomrule
    \end{tabularx}
    \caption{Overview of compute hardware commonly used on autonomous flying vehicles. Due to the emerging trend of deploying learning-based methods onboard, hardware solutions are grouped according to the presence of a general-purpose GPU, enabling real-time inference.}
    \label{tab:hardware}
\end{table*}

\mypara{Low-Level Flight Controller}
\label{sec:lowlevel}
The Low-Level Flight Controller provides real-time low-latency interfacing and control.
A simple and widespread option is the open-source BetaFlight~\cite{Betaflight} software which runs on many commercially available flight controllers, such as the Radix\cite{radix}.
However, BetaFlight is made for human-piloted drones and optimized for a good human flight feeling, but not for autonomous operation.
Furthermore, even though it uses high-speed \ac{IMU} readings for the control loop, it only provides very limited sensor readings at only \SI{10}{\hertz}.
Therefore, Agilicious provides its own low-level flight controller implementation called "\agilow{}", reusing the same hardware as the BetaFlight controllers.
This means that the wide variety of commercially available products can be bought and reflashed with \agilow{} to provide a low-level controller suited for autonomous agile flight.

In particular, we recommend using the BrainFPV Radix~\cite{radix} controller, to deploy our \agilow{} software.
The \agilow{} is based on the open-source NuttX~\cite{NuttX} real-time operating system, optimized to run on embedded microcontrollers such as the STM32F4 used in many BetaFlight products.
Our \agilow{} implementation interfaces with the motors' electronic speed controller (ESC) over the digital bi-directional DShot protocol, allowing not only to command the motors, but also receive individual rotor speed feedback.
This feedback is provided to the high-level controller together with \ac{IMU}, battery voltage, and flight mode information over a \SI{1}{\mega\baud} serial bus at \SI{500}{\hertz}.
The \agilow{} also provides closed-loop motor speed control, bodyrate control, and measurement time synchronization, allowing estimation and control algorithms to take full advantage of the available hardware.

\subsection{Flight Hardware} \label{sec:flight_hardware}

To maximize the agility of the drone, it needs to be designed as lightweight and small as possible~\cite{kumar2012opportunities} while still being able to accommodate the Jetson TX2 compute unit. 
With this goal in mind, we provide a selection of cheap off-the-shelf drone components summarized in Tab.~\ref{tab:hardware}.
The Armattan Chameleon \SI{6}{\inch} frame is used as a base since it is one of the smallest frames with ample space for the compute hardware. 
Being made out of carbon fiber, it is durable and lightweight.
The other structural parts of the quadrotor are custom-designed plastic parts (PLA and TPU material) and produced using a 3D printer. 
Most components are made out of PLA which is stiffer and only parts that act as impact protectors or as predetermined breaking points are made out of TPU.
For propulsion, a \SI{5.1}{\inch} three-bladed propeller is used in combination with a fast-spinning brushless DC motor rated at a high maximum power of \SI{758}{\watt}. 
The chosen motor-propeller combination achieves a continuous static thrust of {4$\times$ \SI{9.5}{\newton}} on the quadrotor and consumes about \SI{400}{\watt} of power per motor.
To match the high power demand of the motors, a lithium-polymer battery with \SI{1800}{\milli\ampere\hour} and a rating of 120C is used.
Therefore, the total peak current of \SI{110}{\ampere} is well within the \SI{216}{\ampere} limit of the battery.
The motors are powered by an electronic speed controller in the form of the Hobbywing XRotor ESC, due to its compact form factor, its high continuous current rating (\SI{60}{\ampere} per motor), and support of the DShot protocol supporting motor speed feedback.

\begin{table}[t]
    \centering
    \footnotesize
    \def\arraystretch{1.2}
    \begin{tabularx}{\linewidth}{l|l|X}
        \toprule
        Component & Product & Specification \\
        \midrule
        Frame & Armattan Chameleon \SI{6}{\inch} & \SI{4}{\milli\meter} carbon fiber, \SI{86}{\gram} \\
        Motor & Xrotor 2306 & 23$\times$\SI{6}{\milli\meter} stator, \SI{2400}{\kilo\volt}, \SI{758}{\watt}, 4$\times$ \SI{27.5}{\gram} \\
        Propeller & Azure Power SFP 5148 & \SI{5.1}{\inch} length and \SI{4.8}{\inch} pitch, 4$\times$ \SI{5}{\gram} \\
        Battery & Tattoo R-Line 1800 & 4$\times$ \SI{3.7}{\volt}, \SI{1800}{\milli\ampere\hour}, \SI{199}{\gram} \\
        Flight Controller & BrainFPV radix & BetaFlight or custom firmware, \SI{6}{\gram} \\
        Motor Controller & HobbyWing XRotor & DShot protocol, 4$\times$ \SI{60}{\ampere}, \SI{15}{\gram} \\
        Compute Unit & nVidia Jetson TX2 & 6$\times$ ARM \SI{2.0}{\giga\hertz}, 256$\times$ CUDA cores, \SI{8}{\giga\byte}, \SI{154}{\gram} \\
        \bottomrule
    \end{tabularx}
    \caption{Overview of the components of the flight hardware design.}
    \label{tab:my_label}
    \vspace*{-12pt}
\end{table}

\rebuttal{
\subsection{Sensors}
To navigate arbitrary uninstrumented environments, drones need means to measure their absolute or relative location and state.
Due to the size and weight constraints of aerial vehicles, and especially the direct impact of weight and inertia on the agility of the vehicle, \acl{VIO} has proven to be the go-to solution for aerial navigation.
The complementary sensing modality of cameras and \ac{IMU}s, their low price and excellent availability, together with the depth-sensing capabilities of stereo camera configurations allow for a simple, compact, and complete perception setup.

Furthermore, our \agilow flight controller already provides high-rate filtered inertial measurements and can be combined with any off-the-shelf camera~\cite{SevenSense,MyntEye,BlueFox} and open-source~\cite{Forster17troSVO,SVO,Qin17tro,Geneva2020icra} or commercial~\cite{SlamCore,ArtiSense} software to implement a \ac{VIO} pipeline.
Additionally, there exist multiple fully integrated products providing out-of-the-box \ac{VIO} solutions, such as the Intel RealSense~\cite{Realsense_data_sheet}, the Roboception rc\_visard~\cite{Roboception}, and the ModalAI Voxl CAM~\cite{ModalAI}.
}

\subsection{The Agilicious Flight Stack Software} \label{sec:flight_stack}
To exploit the full potential of our platform and enable fast prototyping, we provide the Agilicious flight stack as an open-source software package.
The main development goals for Agilicious are aligned with our overall design goals: high versatility and modularity, low latency for agile flight, and transferability between simulation and multiple real-world platforms.
These goals are met by splitting the software stack into two parts.

The core library, called "agilib", is built with minimal dependencies but provides all functionality needed for agile flight, implemented as individual modules (illustrated in Figure~\ref{fig:overview}).
It can be deployed on a large range of computing platforms, from lightweight low-power devices to parallel neural network training farms built on heterogeneous server architectures.
This is enabled by avoiding dependencies on other software components that could introduce compatibility issues and rely only on the core C++-17 standard and the Eigen library for linear algebra.
Additionally, agilib includes a standalone set of unit tests and benchmarks that can be run independently, with minimal dependencies, and in a self-contained manner.

To provide compatibility to existing systems and software, the second component is a ROS-wrapper, called "agiros", which enables networked communication, data logging, provides a simple GUI interface to command the vehicle and allows for integration with other software components.
This abstraction between "agiros" and the core library "agilib" allows a more flexible deployment on systems or in environments where ROS is not available, not needed, or communication overhead must be avoided.
On the other hand, the ROS-enabled Agilicious provides versatility and modularity due to a vast number of open-source ROS packages provided by the research community.

For flexible and fast development, "agilib" uses modular software components unified in an umbrella structure called "pipeline" and orchestrated by a control logic, called "pilot".
The modules consist of an "estimator", "sampler", "controllers", and a "bridge", all working together to track a so-called "reference". 
These modules are executed in sequential order (illustrated in Figure~\ref{fig:overview}) within a forward pass of the pipeline, corresponding to one control cycle.
However, each module can spawn its individual threads to perform parallel tasks, e.g. asynchronous sensor data processing.
Agilicious provides a collection of state-of-the-art implementations for each module inherited from base classes, allowing to create new implementations outside of the core library, and linking them into the pilot at runtime.
Moreover, Agilicious is not only capable to control a drone when running onboard the vehicle, but can also run offboard on computationally more capable hardware and send commands to the drone over low-latency wireless serial interfaces.

Finally, the core library is completed by a physics simulator.
While this might seem redundant due to the vast variety of simulation pipelines available \cite{Furrer2016rotors, shah2018airsim, yunlong2020flightmare}, it allows to use high-fidelity models (e.g. \ac{BEM}~\cite{bauersfeld2021neurobem} for aerodynamics), evaluate software prototypes without having to interface with other frameworks, avoids dependencies, and enables even simulation-based continuous integration testing that can run on literally any platform.
The pilot, software modules, and simulator are all described in the following sections.

\subsubsection{Pilot} \label{sec:pilot}
The pilot contains the main logic needed for flight operation, handling of the individual modules, and interfaces to manage references and task commands.
In its core, it loads and configures the software modules according to YAML~\cite{YAML} parameter files, runs the control loop, and provides simplified user interfaces to manage flight tasks, such as position and velocity control or trajectory tracking.
For all state descriptions, we use a right-handed coordinate system located in the center of gravity, with the $_B\bm{e}_z$ pointing in body-relative upward thrust direction, and $_B\bm{e}_x$ pointing along with the drone's forward direction.
Motion is represented with respect to an inertial world frame with $_I\bm{e}_z$ pointing against the gravity direction, where translational derivatives (e.g. velocity) are expressed in the world frame and rotational derivatives (e.g. bodyrate) are expressed in the body frame.

\mypara{Pipeline}
The pipeline is a distinct configuration of the sequentially processed modules.
These pipeline configurations can be switched at runtime by the pilot or the user, allowing to switch to backup configurations in an emergency, or quickly alternate between different prototyping configurations.

\mypara{Estimator}
The first module in the pipeline is the estimator, which provides a time-stamped state estimate for the subsequent software modules in the control cycle.
A state estimate $\bm{x} = [\bm{p}, \bm{q}, \bm{v}, \bm{\omega}, \bm{a}, \bm{\tau}, \bm{j}, \bm{s}, \bm{b}_\omega, \bm{b}_{a}, \bm{f}_{d}, \bm{f}]$, represents position $\bm{p}$, orientation unit quaternion $\bm{q}$, velocity $\bm{v}$, bodyrate $\bm{\omega}$, linear $\bm{a}$ and angular $\bm{\tau}$ accelerations, jerk $\bm{j}$, snap $\bm{s}$, gyroscope and accelerometer bias $\bm{b}_\omega$ and $\bm{b}_a$, and desired and actual single rotor thrusts $\bm{f}_d$ and $\bm{f}$.
Agilicious provides a feed-through estimator to include external estimates or ground-truth from a simulation, as well as two extended Kalman filters, one with \ac{IMU} filtering, and one using the \ac{IMU} as propagation model.
These estimators can easily be replaced or extended to work with additional measurement sources, such as GPS or altimeters, other estimation systems, or even implement complex localization pipelines such as visual-inertial odometry.

\mypara{Sampler}
For trajectory tracking using a state estimate from the aforementioned estimator, the controller module needs to be provided with a subset of points of the trajectory that encode the desired progress along it, provided by the sampler. 
Agilicious implements two types of samplers: a time-based sampling scheme that computes progress along the trajectory based on the time since trajectory start, and a position-based sampling scheme that selects trajectory progress based on the current position of the platform, trading off temporally accurate tracking for higher robustness and lower positional tracking error.

\mypara{Controller}
\label{sec:controller}
To control the vehicle along the sampled reference setpoints, a multitude of controllers are available, which provide the closed-loop commands for the low-level controller.
We provide a state-of-the-art \ac{MPC} that uses the full non-linear model of the platform and which allows to track highly agile references using single-rotor thrust commands or bodyrate control.
Additionally, we include a cascaded geometric controller based on the quadrotor's differential flatness~\cite{mellinger2011minimum}.
The pipeline can cascade two controllers, which even allows combining the aforementioned \ac{MPC} \cite{Sun2021arxiv} or geometric approaches with an intermediate controller for which we provide an L1 adaptive controller~\cite{Hanover2021icra} and an incremental nonlinear dynamic inversion controller~\cite{Sun2021arxiv}.

\mypara{Bridge}
A bridge serves as an interface to hardware or software by sending control commands to a low-level controller or other means of communication sinks.
Low-level commands can either be single rotor thrusts or bodyrates in combination with a collective thrust.
Agilicious provides a set of bridges to communicate via commonly used protocols such as ROS, SBUS, and serial. 
While the ROS-bridge can be used to easily integrate Agilicious in an existing software stack that relies on ROS, the SBUS protocol is a widely used standard in the \ac{FPV} community and therefore allows to interface Agilicious to off-the-shelf flight controllers such as BetaFlight~\cite{Betaflight}.
For simple simulation, there is a specific bridge to interface with the popular RotorS~\cite{Furrer2016rotors} simulator, which is however less accurate than our own simulation described in Sec.~\ref{sec:simulation}.
As Agilicious is written in a general abstract way, it runs on onboard compute modules and offboard, for which case we provide a bridge to interface with the LAIRD\cite{LAIRD} wireless serial interface.
Finally, Agilicious also provides a bridge to communicate to the custom low-level controller described in Sec.~\ref{sec:lowlevel}.
This provides the advantage of gaining access to closed-loop single rotor speed control, high-frequency \ac{IMU}, rotor speed, and voltage measurements at \SI{500}{\hertz}, all provided to the user through the bridge.

\mypara{References} 
References are used in conjunction with a controller to encode the desired flight path of a quadrotor. 
In Agilicious, a reference is fed to the sampler, which generates a receding-horizon vector of setpoints that are then passed to the controller. 
The software stack implements a set of reference types, consisting of \textit{Hover}, \textit{Velocity}, \textit{Polynomial}, and \textit{Sampled}.
While \textit{Hover} references are uniquely defined by a reference position and a yaw angle, a \textit{Velocity} reference specifies a desired linear velocity with a yaw rate. 
By exploiting the differential flatness of the quadrotor platform, \textit{Polynomial} references describe the position and yaw of the quadrotor as polynomial functions of time. 
Sampled references provide the most general reference representations.
Agilicious provides interfaces to generate, and receive such sampled references and also defines a message and file format to store references to a file. 
By defining such formats, a wide variety of trajectories can be generated, communicated, saved, and executed using Python or other languages.
Finally, to simplify the integration and deployment of other control approaches, Agilicious also exposes a command feedthrough, that allows taking direct control over the applied low-level commands.
For safety, even when command feedthrough is used, Agilicious provides readily available back-up control that can take over on user request or on timeout.

\rebuttal{
\mypara{Guard}
To further support users in fast prototyping, Agilicious provides a so-called guard.
This guard uses the quadrotor's state-estimate or an alternative estimate (e.g. from motion capture when flying with \ac{VIO} prototypes) together with a user-defined spatial bounding box to detect unexpected deviations from the planned flight path.
Further detection metrics can be implemented by the user.
Upon violation of e.g. the spatial bounding box, the guard can switch control to an alternative pipeline using a backup estimate and control configuration.
This safety pipeline can e.g. use a motion capture system and a simple geometric controller, while the main pipeline runs a \ac{VIO} estimator, an \ac{MPC}, reinforcement learning control strategies, or other software prototypes.
By providing this measure of backup, Agilicious significantly reduces the risk of crashes when testing novel algorithms, and allows to iterate over research prototypes faster.
}

\subsubsection{Simulation} \label{sec:simulation}
The Agilicious software stack includes a simulator that allows simulating quadrotor dynamics at various levels of fidelity to accelerate prototyping and testing. 
Specifically, Agilicious models  motor dynamics and aerodynamics acting on the platform. To also incorporate the different, possibly off-the-shelf, low-level controllers that can be used on the quadrotors, the simulator can optionally simulate the behavior of low-level controllers.
One simulator update, typically called at \SI{1}{\kilo\hertz}, includes a call to the simulated low-level controller, the motor model, the aerodynamics model, and the rigid body dynamics model in a sequential fashion. 
Each of these components is explained in the following. 

\mypara{Low-Level Controller \& Motor Model} Simulated low-level controllers run at simulation frequency and convert collective thrust and bodyrate commands into individual motor speed commands. The usage of a simulated low-level controller is optional if the computed control commands are already in the form of individual rotor thrusts. In this case, the thrusts are mapped to motor speed commands and then directly fed to the simulated motor model.
The motors are modeled as a first-order system with a time constant which can be identified on a thrust test stand.

\mypara{Aerodynamics}
The simulated aerodynamics model lift and drag produced by the rotors from the current ego-motion of the platform and the individual rotor speeds. 
Agilicious implements two rotor models: \textit{Quadratic} and \textit{BEM}. 
The \textit{Quadratic} model implements a simple quadratic mapping from rotor rotational speed to produced thrust, as commonly done in quadrotor simulators~\cite{Furrer2016rotors, shah2018airsim, yunlong2020flightmare}. 
While such a model does not account for effects imposed by the movement of a rotor through the air, it is highly efficient to compute. 
In contrast, the \textit{BEM} model leverages Blade-Element-Momentum-Theory~(BEM) to account for the effects of varying relative airspeed on the rotor thrust. To further increase the fidelity of the simulation, a neural network predicting the residual forces and torques (e.g. unmodeled rotor to rotor interactions and turbulence) can be integrated into the aerodynamics model. For details regarding the \textit{BEM} model and the neural network augmentation, we refer the reader to~\cite{bauersfeld2021neurobem}. 

\mypara{Rigid Body Dynamics}
Provided with a model of the forces and torques acting on the platform predicted by the aerodynamics model, the system dynamics of the quadrotor are integrated using a 4th order Runge-Kutta scheme with a step size of \SI{1}{\milli\second}. 
Agilicious also implements different integrators such as explicit Euler or symplectic Euler. 

Apart from providing its own state-of-the-art quadrotor simulator, Agilicious can also be interfaced with external simulators. 
Interfaces to the widely-used RotorS quadrotor simulator~\cite{Furrer2016rotors} and Flightmare~\cite{yunlong2020flightmare}, including the \ac{HIL} simulator, are already provided in the software stack.

\section{Acknowledgments}
\mypara{Author Contribution}
P.F. developed the Agilicious software concepts and architecture, contributed to the Agilicious implementation, helped with the experiments, and wrote the manuscript.
E.K. contributed to the Agilicious implementation, designed the Agilicious hardware, helped with the experiments, and wrote the manuscript.
A.R., R.P., S.S., and L.B. contributed to the Agilicious implementation, helped with the experiments, and wrote the manuscript.
T.L. evaluated the hardware components, designed and built the Agilicious hardware, helped with the experiments, and wrote the manuscript.
Y.S. contributed to the Agilicious implementation, helped with the experiments, and wrote the manuscript.
A.L. helped with the experiments and wrote the manuscript.
D.S. provided funding, contributed to the design and analysis of the experiments, and revised the manuscript.

\mypara{Funding}
This work was supported by the National Centre of Competence in Research (NCCR) Robotics through the Swiss National Science Foundation (SNSF) and the European Union’s Horizon 2020 Research and Innovation Program under grant agreement No. 871479 (AERIAL-CORE) and the European Research Council (ERC) under grant agreement No. 864042 (AGILEFLIGHT).

\mypara{Data and Materials}
The main purpose of this paper is to share our data and materials. Therefore all materials, both software as well as hardware designs, are open-source accessible at {\small\url{https://agilicious.dev}} under the GPL v3.0 license. 

\balance
\bibliography{bibliography}

\end{document}